\documentclass[runningheads]{llncs}
\usepackage{graphicx}
\usepackage{graphicx}

\usepackage{times}
\usepackage{latexsym}

\usepackage{times}
\usepackage{latexsym}
\usepackage{cite}
\usepackage{amsmath,amssymb,amsfonts}
\usepackage{algorithmic}
\usepackage{graphicx}
\usepackage{textcomp}
\usepackage{xcolor}
\usepackage{array}
\usepackage[linesnumbered,ruled]{algorithm2e}
\usepackage{enumerate}
\usepackage{url}
\usepackage{multirow}
\usepackage{booktabs}
\usepackage{url}
\usepackage{caption} 
\begin{document}
\title{Disentangling Latent Emotions of Word Embeddings on Complex Emotional Narratives}
%
%
\author{Zhengxuan Wu\inst{1}\orcidID{0000-0001-5581-8908} \and
Yueyi Jiang\inst{2}\orcidID{0000-0002-5311-4648}}
\authorrunning{Z. Wu, Y. Jiang}
%
\institute{Stanford University, Stanford CA 94085, USA\\
\email{wuzhengx@stanford.edu}
\and
University of California San Diego, La Jolla CA 92093, USA\\
\email{yujiang@ucsd.edu}
}
\maketitle              
\begin{abstract}
Word embedding models such as GloVe are widely used in natural language processing (NLP) research to convert words into vectors. Here, we provide a preliminary guide to probe latent emotions in text through GloVe word vectors. First, we trained a neural network model to predict continuous emotion valence ratings by taking linguistic inputs from Stanford Emotional Narratives Dataset (SEND). After interpreting the weights in the model, we found that only a few dimensions of the word vectors contributed to expressing emotions in text, and words were clustered on the basis of their emotional polarities. Furthermore, we performed a linear transformation that projected high dimensional embedded vectors into an \textit{emotion space}. Based on NRC Emotion Lexicon (EmoLex), we visualized the entanglement of emotions in the lexicon by using both projected and raw GloVe word vectors. We showed that, in the proposed \textit{emotion space}, we were able to better disentangle emotions than using raw GloVe vectors alone. In addition, we found that the sum vectors of different pairs of emotion words successfully captured expressed human feelings in the EmoLex. For example, the sum of two embedded word vectors expressing \textit{Joy} and \textit{Trust} which express \textit{Love} shared high similarity (similarity score .62) with the embedded vector expressing \textit{Optimism}. On the contrary, this sum vector was dissimilar (similarity score -.19) with the the embedded vector expressing \textit{Remorse}. In this paper, we argue that through the proposed \textit{emotion space}, arithmetic of emotions is preserved in the word vectors. The affective representation uncovered in emotion vector space could shed some light on how to help machines to disentangle emotion expressed in word embeddings.
\end{abstract}
\section{Introduction}
Constructing human-friendly Artificial Intelligence (AI) is essential for humans as it will help us get the most benefits from AI systems~\cite{ong2015affective}. Being able to detect emotions through language is the building block of such AI agents~\cite{preston2002empathy, morelli2017empathy}. One such way is through word embeddings - encoding words in vectors. Researchers have proposed various word embedding methods such as GloVe and Word2Vec~\cite{pennington2014glove, mikolov2013efficient}. However, to date, understanding the expressed human emotions in text from word embeddings by an agent remains a challenging problem, as word embedding based models are generally missing the direct interpretations of the word vectors~\cite{gu2017non, bordes2014open}. 

In this study, we provide different strategies for interpreting the emotional semantics of words through word embeddings. We visualize word clusters by projecting word vectors into 2-dimensional space where embedding vectors are clustered by their emotional polarities. Additionally, based on the weights of a pretrained neural network model, we are able to project words into an \textit{emotion space}. we show that the arithmetic of emotions holds, which is consistent with the principle introduced by Plutchik~\cite{plutchik1980general}. An example is as follows:
\begin{align}
    v_{\text{Love}} & = v_{\text{Joy}} + v_{\text{Trust}}
\end{align}
We also show that words with opposite emotion valence separated in the \textit{emotion space}. Rather than relying on dictionaries or hidden layers of neural networks, we provide a preliminary method of probing emotion entanglements in word vectors, making one initial step in exploring the latent emotions in word embeddings from modeling emotions in complex narratives.
\section{Related Works}
Word embeddings are widely applied in sentiment analysis with neural network models~\cite{ ebrahimi2015recurrent, donahue2015long, wu2019attending}. However, these models often lack clear interpretations of word vectors~\cite{li2015visualizing}. To date, only few studies have probed the semantics of emotions from word embeddings~\cite{seyeditabari2017can, li2017inferring}. These studies attempted to interpret natural language models through visualization of word vectors and hidden layers of the models. For example, researchers have visualized the hidden layers' representation of word vectors in 2-dimensional space in which words with similar meanings are clustered together~\cite{ji2014representation, faruqui2014improving}. Additionally, other methods have focused on visualizing the hidden layers of neural network models using gradients and weights inferred from the models~\cite{li2015visualizing, seyeditabari2017can, li2017inferring}.


In this study, we aim to provide a systematic way of identifying emotions directly from text. Using embedded vectors, our method is different from the existing research that has focused on deriving latent semantic information from hidden layers of the neural network models ~\cite{li2015visualizing, li2017inferring}. Throughout the paper, we provide preliminary evidence for detecting emotions in word vectors through word embeddings specifically in emotional expressions.

\section{Dataset}
In this paper, we used Stanford Emotional Narratives Dataset (SEND) as our dataset. SEND is comprised of transcripts of video recordings in which participants shared emotional stories, and it has been well explored in computational models of emotion~\cite{ong2019SEND, wu2019attending}. In each transcript, timestamps were generated for every word based on force-alignments\footnote{https://github.com/ucbvislab/p2fa-vislab} of audio inputs, and continuous emotional valence ratings were collected by annotators\footnote{We have 25 ratings per transcript from annotators. The target variable is the average collected ratings.}. These ratings serve as the target variable in our model, which were scaled between [-1,1] and sampled every 0.5s. 

The dataset includes 193 transcripts that last on average 2 mins 15 secs, for a total duration of 7 hrs and 15 mins. We divided these transcripts into a \textbf{Train set} (60\% of the dataset, 117 videos, 38 targets, 4 hrs 26 mins long), a \textbf{Validation set} (20\%, 38 videos, 27 targets, 1 hr 23 mins long) and a \textbf{Test set} (20\%, 38 videos, 27 targets, 1 hr 26 mins long).

\section{Autoregressive Model}
To interpret the word vectors, we trained an autoregressive linear model to predict emotional valence ratings. We first used 300-dimensional GloVe word vectors which were pre-trained on wikipages~\cite{pennington2014glove}. Then, we used \texttt{interp} function in \texttt{numpy} package to assign each word a valence rating by linearly interpolating the original ratings using the timestamps for each word. 

By concatenating the word vector $v_t \in \mathbb{R}^{e\times 1}$ where $e$ is 300 for GloVe from the current time point (at time $t$) with the hidden state vector $h_{t-\tau} \in \mathbb{R}^{300\times 1}$ where $h$ is 300 from last time point (at time $t-\tau$), we produced a hidden vector $h_{t} \in \mathbb{R}^{600\times 1}$ for the current time point. The hidden vector was then passed into a linear layer with bias to produce a output vector $o_{t} \in \mathbb{R}^{300\times 1}$. Subsequently, the output vector was passed into a linear layer which produced a single pseudo-rating prediction $u_t$ for the current time point. The final rating prediction $r_t$ was produced by a self-learned linear filter by taking $r_{t-\tau}$ into account:
\begin{align}
    h_t & = \text{Concat}(h_{t-\tau}, v_t) \\
    o_{t} &= [\textbf{W}_{\text{h}}, \textbf{W}_{\text{v}}] h_t + \textbf{b}_{\text{h}} \\
    u_t &= \textbf{W}_{\text{o}}o_{t} + \textbf{b}_{\text{o}} \\
    r_t &= \sigma r_{t-\tau} + (1-\sigma) u_t
\end{align}
with weight matrices $\textbf{W}_{\text{h}}, \textbf{W}_{\text{v}} \in \mathbb{R}^{300\times 300}$, $\textbf{W}_{\text{o}} \in \mathbb{R}^{300\times 1}$ and bias vectors $\textbf{b}_{\text{h}} \in \mathbb{R}^{600\times 300}$, $\textbf{b}_{\text{o}} \in \mathbb{R}^{300\times 1}$. We used $\sigma$ to denote the weight on previous rating prediction.

\section{Model Evaluation}
\begin{figure*}[!bt]
\centering
\includegraphics[width=1\textwidth]{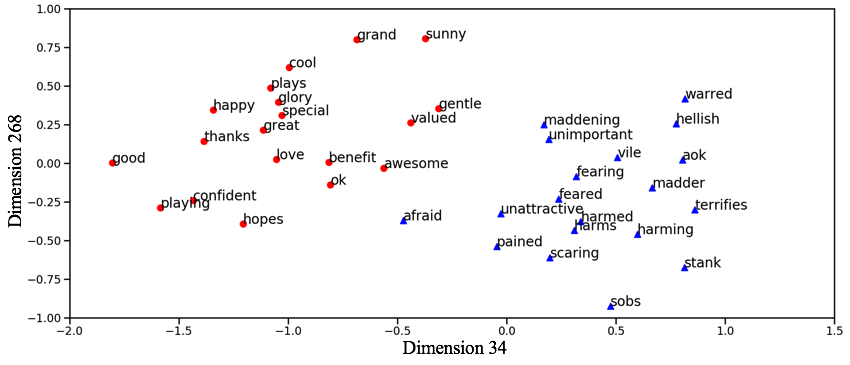}
\caption{Visualization of word clusters by their emotional polarities (i.e., positive or negative). Positive words are represented as red circles, while negative words are shown in blue triangles.}
\label{fig:2dspace}
\end{figure*}
Before interpreting the model results, we ensured that the optimal performance was achieved. Similar to the evaluation metric used in a previous study on SEND~\cite{wu2019attending}, Concordance Correlation Coefficient (CCC, as defined by \cite{lin1989concordance}) was evaluated. Specifically, we compared model performance on the \textbf{Validation set} and \textbf{Test set} with the human benchmark provided by SEND. Our model achieved a CCC of $.37 \pm .11$ on the \textbf{Validation set} and $.35 \pm .15$ on the \textbf{Test set}, comparing to human's performance of a CCC of $.47 \pm .12$ on the \textbf{Validation set} and $.46 \pm .14$ on the \textbf{Test set}. In our final model, the learnable parameter $\sigma$ is 0.84, indicating that the current prediction at time $t$ is primarily dependent on the previous state at time $t-1$.

\section{Experiment And Results}
\subsection{Pure On Weights}
By using the weights from two linear layers, each dimension was assigned a score to quantify its contribution to emotion valence in words (Alg.1). Higher absolute weights are associated with larger gradient changes in outputs, indicating higher importance in emotion expression in a given dimension.
\begin{algorithm}
\SetAlgoLined
\KwResult{Scores for each dimension}
 dim\_scores = \{\}\;
 \For{$i \in \{1,...,300\}$}{
  \_score = 0.0\;
  \For{$w_{j} \in \textbf{W}_{\text{v}_{i}}$}{
    \_score += $w_{j} \cdot \textbf{W}_{\text{o}_{i}}$\;
  }
  dim\_scores[$i$] = abs(\_score)\;
 }
    \caption{Scoring algorithm for ranking important dimensions of GloVe vectors in emotion expression}
\end{algorithm}

\begin{figure}[!bt]
\centering
\includegraphics[width=\columnwidth]{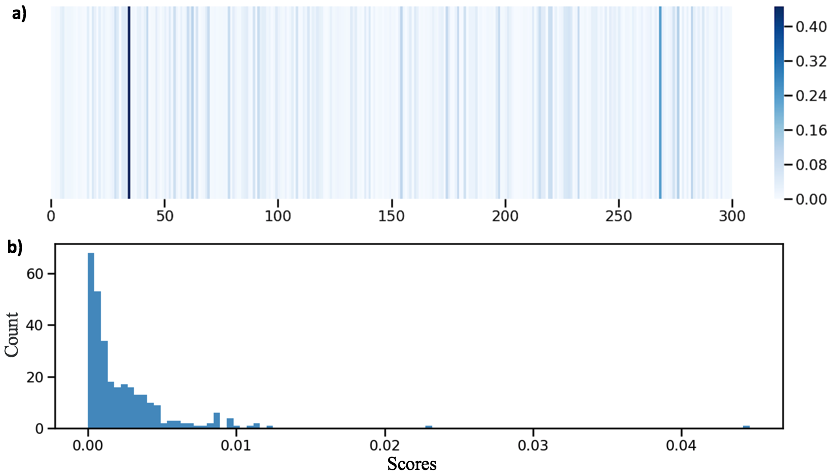}
\caption{Illustration of calculated scores on each dimension of the 300-dimensional GloVe vectors. (a) Heatmap of scores for each dimension in GloVe vectors. (b) Distribution of scores across all dimensions.}
\label{fig:Heatmap}
\end{figure}
Based on the scores, we first produced a heatmap (Fig.\ref{fig:Heatmap}.a) for all 300 dimensions in GloVe vectors to visualize the importance of each dimension. In addition, we plotted the distribution of scores (Fig.\ref{fig:Heatmap}.b). We found that a large portion of dimensions were non-expressive in emotion valence. Specifically, we discovered that the $34^{th}$ dimension of the GloVe vectors is the most important dimension in expressing emotions in word vectors.

\subsection{2-d Emotion Visualization}

Based on the scores for all dimensions, we picked out the top 2 dimensions and visualized the clustering effect of words. We used out-of-sample words from LIWC 2007 ~\cite{pennebaker2001linguistic} from which we selected top 19 words ranked by their gradients of forward propagation for positive and negative polarities, respectively. Figure \ref{fig:2dspace} shows that words with positive meaning are well separated from words with negative meaning in this space. Subsequently, we showed that the dimensions picked by our scoring algorithm (Alg.1) could be used to separate words into clusters that represent two emotion polarities, positive and negative emotions.

\subsection{Entanglement Of Emotions}
\begin{figure}[!bt]
\centering
\includegraphics[width=\columnwidth]{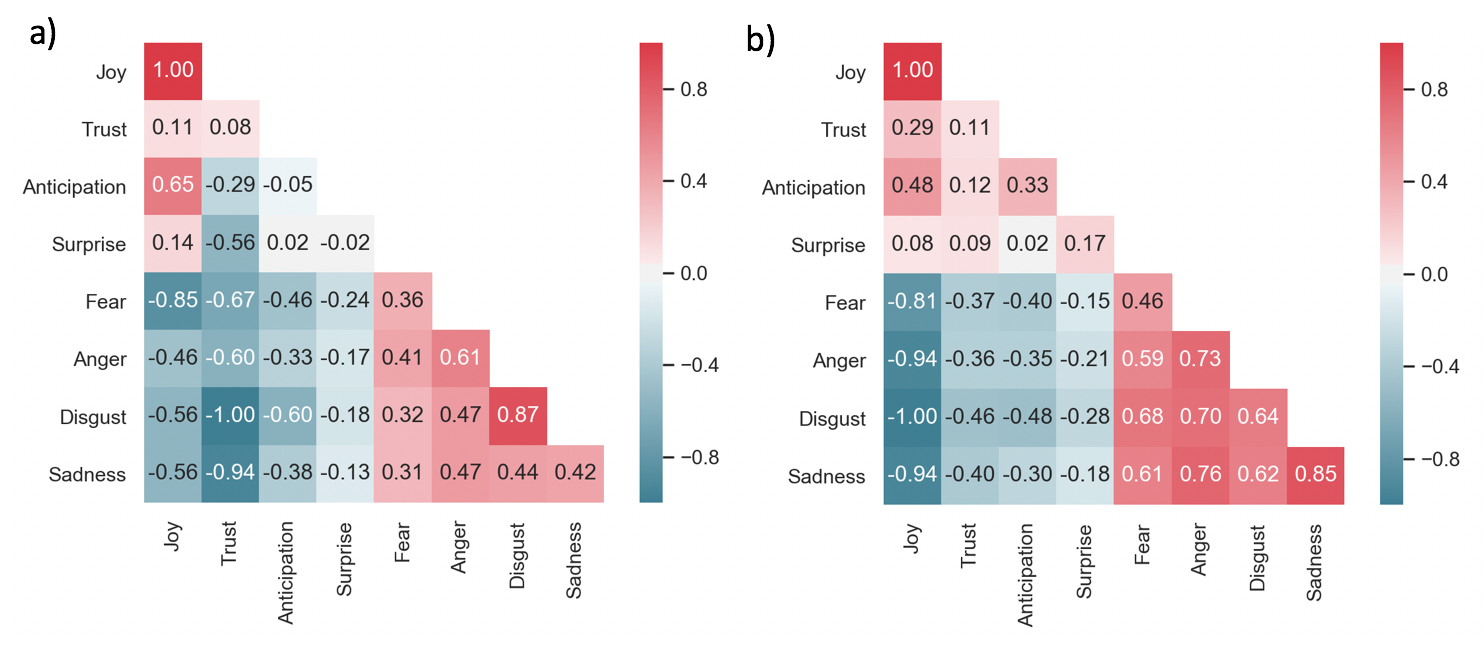}
\caption{Heatmaps of cosine similarities scores between words with paired emotions. (a) is produced by using the raw GloVe vector. (b) is produced by using the projected GloVe vectors.}
\label{fig:EmoHeat}
\end{figure}
EmoLex has eight categories of word groups by emotions: \textit{Joy}, \textit{Trust}, \textit{Anticipation}, \textit{Surprise}, \textit{Fear}, \textit{Anger}, \textit{Disgust} and \textit{Sadness}. To show the entanglements of emotions in GloVe word vectors, we first randomly selected word pairs from any two distinct emotions of the eight emotion categories. We then exhaustively calculated the average cosine similarity scores between these word pairs and produced heatmaps with $8\times 8$ similarity scores by
\begin{equation}
{\displaystyle {\text{similarity}}=\cos(\theta )={\mathbf {A} \cdot \mathbf {B}  \over \|\mathbf {A} \|\|\mathbf {B} \|}={\frac {\sum \limits _{i=1}^{n}{A_{i}B_{i}}}{{\sqrt {\sum \limits _{i=1}^{n}{A_{i}^{2}}}}{\sqrt {\sum \limits _{i=1}^{n}{B_{i}^{2}}}}}}}
\end{equation}

To investigate if words are better clustered by emotional polarity in the proposed $\textit{emotion space}$, we projected word vectors to this space by calculating element-wise multiplication of the weight $\textbf{W}_{\text{v}}$ from our model and raw word vectors. Based on the heatmap (Fig. \ref{fig:EmoHeat}), we found that word vectors in the \textit{emotion space} were better clustered by emotional polarity than raw GloVe vectors. For example, using the raw GloVe vectors, words expressing \textit{Anticipation} had a low similarity score (-.29) with \textit{Trust}, even though both words are associated with positive valence. However, after we projected the words into the \textit{emotion space}, the similarity scores drastically increased. Thus, the results suggest that the projected GloVe vectors may provide better interpretations of emotion expressions in words.

\subsection{Arithmetic Of Emotions}
\begin{figure}[!bt]
\centering
\includegraphics[width=\columnwidth]{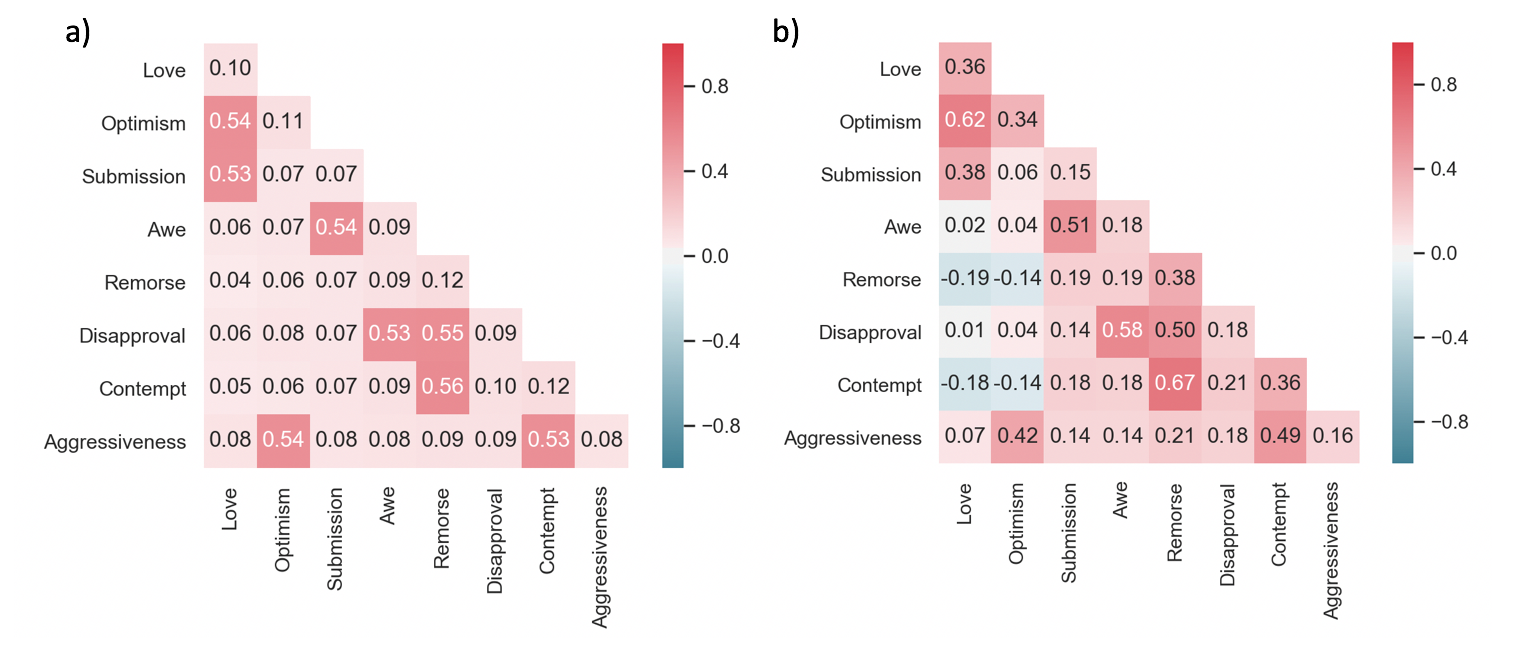}
\caption{Heatmaps of cosine similarities scores between words with paired feelings. (a) is produced by using the raw GloVe vector. (b) is produced by using the projected GloVe vectors.}
\label{fig:FeelingHeatmap}
\end{figure}
\begin{table*}[t]
\begin{center}
  \begin{tabular}{ c | c | c | c }
    \hline
    \textbf{Feelings} & \textbf{Emotions} & \textbf{Opposite} & \textbf{Emotions} \\ \hline
    Love & Joy + Trust & Remorse & Sadness + Disgust \\ \hline
    Optimism & Anticipation + Joy & Disapproval & Surprise + Sadness \\ \hline
    Submission & Trust + Fear & Contempt & Disgust + Anger \\ \hline
    Awe & Fear + Surprise & Aggressiveness & Anger + Anticipation \\
    \hline
  \end{tabular}
  \caption{Taxonomy of feelings and arithmetic of emotions. The \textbf{Opposite} column represents a list of opposite feelings of the first column which is based on Plutchik’s wheel of emotions~\cite{plutchik1980general}.}
\end{center}
\label{tab:arith}
\end{table*}
In this section, we demonstrate that with the linear projection matrix $\textbf{W}_{\text{v}}$ from the pretrained model, word vectors is transformed into a \textit{emotion space} where the arithmetic of emotions (Tab. 1) is better represented compared to using raw GloVe vectors. According to Plutchik’s wheel of emotions~\cite{plutchik1980general}, feelings are combinations of two emotions. For example, \textit{Love} is a combination of \textit{Joy} and \textit{Trust}. We want to examine if this arithmetic of emotions is preserved with word embeddings, which means whether the sum of word vectors expressing \textit{Joy} and \textit{Trust} has a high similarity score with the word vector of \textit{Love} . First, we formulated vectors representing the feeling that EmoLex is missing. For instance, we randomly paired words expressing \textit{Joy} and \textit{Trust} and added up word vectors from each pair to formulate a groups of vectors representing the feeling \textit{Love}. Then, similar to the previous analysis, we calculated the average similarity scores between any two pairs of feelings using the generated word vectors. Ideally, the similarity score between two opposite feelings should be low whereas the score between two similar feelings should be high.

Based on our heatmap (Fig. \ref{fig:FeelingHeatmap}), we found that arithmetic of emotions was not well preserved with raw GloVe vectors given the fact that the vectors of feelings had extremely low similarity scores with themselves. For example, the similarity scores between \textit{Love} and itself is only .10. Meanwhile, opposite feelings had higher similarity scores than expected. In the \textit{emotion space}, the distribution of similarity scores were more systematic. For example, the similarity scores between \textit{Love} and itself increased to .36 whereas the similarity scores between two opposite feelings \textit{Love} and \textit{Remorse} decreased to -.19.

\section{Conclusion}
In the present study, we demonstrate that word embeddings with GloVe preserve latent emotions in text. By ranking weights across dimensions of the GloVe vectors, we show that majorities of dimensions are not associated with emotion representations in words. Additionally, from the top two dimensions ranked by importance, we demonstrate that words can be clustered by emotional polarity (Fig.\ref{fig:2dspace}). 

Using a projection matrix to transform the original GloVe vectors, we find that in the proposed \textit{emotion space}, arithmetic of emotions is better represented compared to using the raw vectors alone. By comparing the similarities across vectors, we demonstrate that words with opposite emotional meanings are well separated in the \textit{emotion space}. Meanwhile, we show that arithmetic of emotions is a good proxy of human feelings in the proposed space, consistent with the Plutchik's theory of emotions. Our preliminary exploration shed some lights on modeling the inter-relations in different emotion categories through word embeddings, and encourages more refined research in probing emotion expressions in other types of word embeddings.

%
%
%
\bibliographystyle{splncs04}
\bibliography{mybibliography}
\end{document}